%% file: main.tex
\documentclass[10pt,letterpaper]{article}
\usepackage{arxiv}
\input{preamble}

\renewcommand{\shorttitle}{InertiaMPDD}
\renewcommand{\headeright}{Preprint}
\title{\normalfont\bfseries InertiaMPDD: Multimodal Personalized Depression Detection via Emotional Inertia-Inspired Dynamics Modeling}
\author{\normalfont\small
Changzeng Fu\textsuperscript{1,2}, Songyang Wang\textsuperscript{2}, Chang Gao\textsuperscript{1}, Xinyi Zhang\textsuperscript{1}\\[3pt]
Peilin Li\textsuperscript{1}, Mo Meng\textsuperscript{1}, Jiarui Wu\textsuperscript{1}, Mingze Li\textsuperscript{1}\\[3pt]
Hongwei Ding\textsuperscript{1}, Shiqi Zhao\textsuperscript{1}, Yong Li\textsuperscript{4}, Lang He\textsuperscript{3}\\[9pt]
\textsuperscript{1}Northeastern University\\[2pt]
\textsuperscript{2}The University of Osaka\\[2pt]
\textsuperscript{3}Xi'an University of Posts and Telecommunications\\[2pt]
\textsuperscript{4}Southeast University\\[5pt]
\texttt{fuchangzeng@qhd.neu.edu.cn}
}
\date{}
\begin{document}
\maketitle
\input{sec/paper}
\FloatBarrier
{\small
\bibliographystyle{ieeenat_fullname_citeorder}
\bibliography{references}
}
\end{document}

%% file: preamble.tex
\usepackage{colortbl}
\definecolor{tableheader}{RGB}{243,245,247}
\definecolor{methodrow}{RGB}{232,241,248}
\usepackage[T1]{fontenc}
\usepackage[utf8]{inputenc}
\usepackage{microtype}
\usepackage{graphicx,booktabs,multirow,array}
\usepackage{amsmath,amssymb,amsthm,mathtools,bm}
\usepackage{placeins,needspace}
\usepackage{algorithm,algorithmic}
\usepackage{adjustbox}
\usepackage[numbers,sort&compress]{natbib}
\usepackage{xcolor}
\usepackage{caption}
\definecolor{linkblue}{rgb}{0.21,0.40,0.60}
\usepackage[breaklinks,colorlinks,allcolors=linkblue]{hyperref}
\usepackage[capitalize,noabbrev]{cleveref}

\newtheorem{proposition}{Proposition}
\newcommand{\sg}{\operatorname{sg}}
\newcommand{\forcear}{\mathrm{force}_{\mathrm{AR}}}
\hypersetup{pdftitle={InertiaMPDD: Multimodal Personalized Depression Detection via Emotional Inertia-Inspired Dynamics Modeling},pdfauthor={Changzeng Fu, Songyang Wang, Chang Gao, Xinyi Zhang, Peilin Li, Mo Meng, Jiarui Wu, Mingze Li, Hongwei Ding, Shiqi Zhao, Yong Li, Lang He},pdfsubject={Research preprint}}

\makeatletter
\@ifundefined{captionof}{\usepackage{capt-of}}{}
\makeatother

%% file: sec/paper.tex
\begin{abstract}
Multimodal depression screening estimates risk from speech, visual behavior and language. Current methods mainly predict labels, with limited evidence of departures from a non-depressed reference. Recording-level supervision also leaves fine-grained emotional dynamics unannotated. Motivated by \mbox{\textbf{emotional inertia}}, the persistence of affect, InertiaMPDD learns healthy next-state patterns using a \textbf{world-model-inspired latent predictor}. Using pretrained multimodal representations, it predicts from observed history and context. Reference deviation, residual decay rate and a Jacobian spectral descriptor form \textbf{reference-derived emotional dynamics descriptors}, evaluated as candidate proxy biomarkers for standalone screening and personalized decision correction. Across five cohorts from four datasets, fusion reaches $98.41$\% F1 on D-Vlog and $84.98$\% on LMVD. Standalone descriptors outperform fixed additive fusion on MPDD but discriminate weakly on AVEC2014. These findings distinguish reference informativeness from fusion benefit and reveal context-dependent utility. An interactive demo is available at \url{https://hacilab.github.io/InertiaMPDD-demo/}.
\end{abstract}

\section{Introduction}
\label{sec:introduction}

Multimodal depression detection uses language, speech and facial behavior to estimate risk from long recordings. Recent sequence models and multimodal large models learn increasingly effective relationships between these observations and depression labels \citep{mmfformer2025,dynmultidep2026,sadeghi2024llm,wang2025screening}. Some also generate symptom-based rationales \citep{depressionagent2026,red2025}. Affective science offers a complementary temporal perspective. \textbf{Emotional inertia describes the tendency of affective states to persist over time}, commonly studied through their lagged autocorrelation and associated with depression \citep{kuppens2010inertia,koval2013affect,houben2015meta}. This motivates looking for measurable dynamical evidence alongside a recording-level risk prediction.

\begin{figure}[!t]
  \centering
  \includegraphics[width=0.68\linewidth]{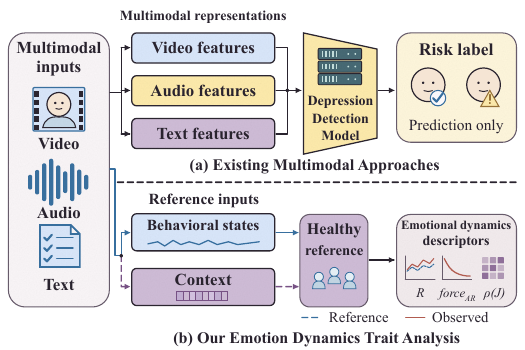}
  \caption{\textbf{Motivation.} (a) Existing Multimodal Approaches. (b) Our Emotion Dynamics Trait Analysis.}
  \label{fig:paradigm}
\end{figure}

Two limitations make this evidence difficult to obtain. \textbf{Coarse supervision leaves emotional dynamics unobserved}: depression datasets commonly assign one label to an entire recording or participant \citep{yoon2022dvlog,lmvd2025,fu2025first}, without identifying the informative emotional segments. Emotion benchmarks such as IEMOCAP instead annotate individual utterances \citep{busso2008iemocap}. Comparable supervision for directly tracking emotional persistence is usually unavailable in depression recordings. \textbf{A population prediction also lacks an explicit behavioral reference}. Slow speech or limited facial movement may be habitual for one speaker and unusual for another. A useful comparison therefore needs to accommodate context while remaining learnable without segment-level emotion or personality labels (\Cref{fig:paradigm}).

Motivated by these limitations, we ask \textbf{whether a pretrained multimodal model can provide measurable evidence of behavioral departures from a non-depressed reference, and whether this evidence can support depression screening}. World models and latent predictive architectures learn representations and predict how observations evolve \citep{ha2018worldmodels,lecun2022path,assran2023ijepa,assran2025vjepa2}. We use this principle to learn a healthy next-state predictor from frozen pretrained representations. It predicts from observed history and context. Comparison with the next observation gives a time-resolved residual. Inspired by emotional inertia, we examine reference deviation, residual persistence and the predictor's local spectral structure. We call these \textbf{reference-derived emotional dynamics descriptors}: proxies computed from behavioral representations, rather than direct measurements of annotated emotions. Their association with depression labels tests their value as candidate proxy biomarkers. Experiments across recording settings assess their screening utility and applicability.

We address this question with InertiaMPDD. Our contributions are:
\begin{itemize}
  \setlength{\itemsep}{2pt}
  \setlength{\parskip}{0pt}
  \setlength{\parsep}{0pt}
  \item We formulate \textbf{reference-based behavioral evidence} under recording-level depression supervision, using emotional inertia to motivate measurable descriptions of predictive departure and persistence.
  \item We introduce the \textbf{Idiographic Inertia Pathway (IIP)}, a world-model-inspired healthy predictor that extracts reference deviation, residual decay rate and a Jacobian spectral descriptor. These descriptors support standalone screening and an individual correction to the multimodal Nomothetic Screening Pathway (NSP).
  \item We evaluate the pathways on \textbf{five cohorts from four datasets}, finding complementary evidence on spontaneous vlogs, stronger standalone IIP than fixed fusion on MPDD and limited reference discrimination on AVEC2014 \citep{yoon2022dvlog,lmvd2025,valstar2014avec,fu2025first}. Feature controls and visualizations distinguish reference informativeness from fusion benefit.
\end{itemize}

\FloatBarrier

\input{sec/related_work}

\section{Method}
\label{sec:method}
\label{sec:overview}

\subsection{Problem statement}
\label{sec:problem}
We observe recordings $\mathcal X_i=(A_i,V_i,L_i)$ containing audio, video and transcript, with recording-level depression labels $y_i\in\{0,1\}$. Let $\mathcal D=\{(\mathcal X_i,y_i)\}_{i=1}^{N}$ denote labeled recordings and $\mathcal D_{\mathrm{train}}$ their training subset. Segment-level emotion labels are unavailable. Our goal is to predict risk while identifying measurable departures from a non-depressed behavioral reference. NSP maps the multimodal recording to a population margin $m_i=M_\phi(\mathcal X_i)$. IIP maps its frozen feature sequence $x_{1:T_i}$ to a reference-derived descriptor vector $u_i=U(x_{1:T_i})$. A supervised head converts $u_i$ into an individual correction $b_i$, giving the screening score $m_i+b_i$.

\begin{figure}[t]
  \centering
  \includegraphics[width=\linewidth]{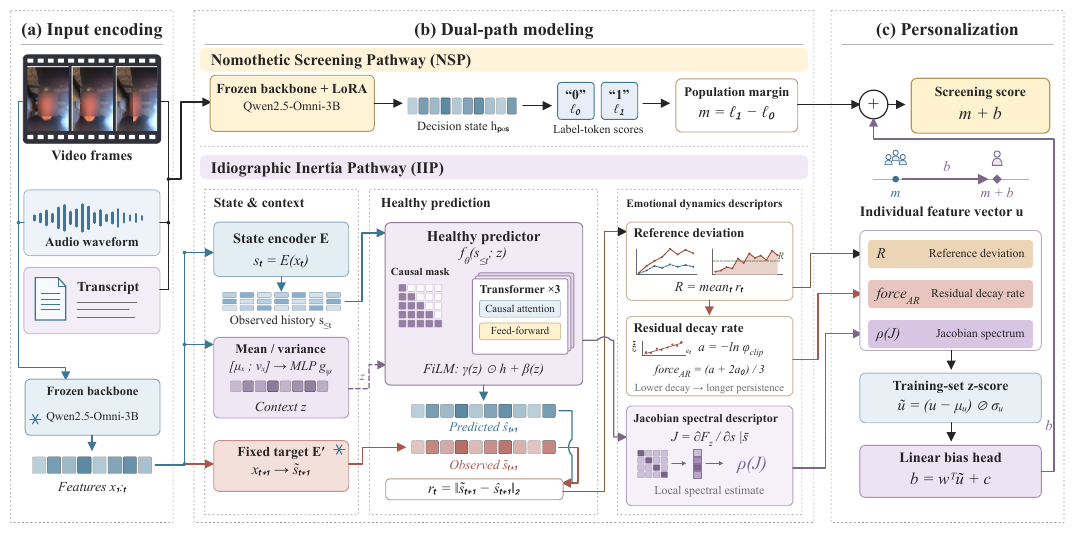}
  \caption{\textbf{InertiaMPDD framework.} (a) Input encoding. (b) NSP population margin and IIP healthy-reference descriptors. (c) Individual correction and screening. Snowflakes mark frozen parameters. Curves and representations are schematic.}
  \label{fig:framework}
\end{figure}

To construct $U$, we learn next-state prediction from non-depressed training recordings. The online state $s_t=E(x_t)$ and context $z$ determine the prediction $\hat s_{t+1}$, which is compared with the independently encoded observation $\tilde s_{t+1}$. This comparison yields recording-level descriptors without assigning emotions to individual segments. Here $z$ conditions the reference, whereas the training-standardized descriptor vector $\tilde u$ adjusts the decision. \Cref{fig:framework} shows this process.

\subsection{Subject-conditioned healthy-reference modeling}
\label{sec:latent}

The frozen backbone provides 2048-dimensional audio and visual features, aligned at $1$~Hz and concatenated into $x_t\in\mathbb{R}^{4096}$. The online encoder produces $s_t=E(x_t)\in\mathbb{R}^{256}$, a learned audiovisual representation rather than an emotion annotation. Audio tokens are pooled over one-second intervals. 64 sampled visual frames are expanded to the same temporal grid by nearest-neighbor assignment. This gives an audiovisual state sequence for each recording.

Context is encoded from the temporal mean $\mu_x$ and elementwise variance $v_x$ of the backbone features. The concatenated statistics give
\begin{equation}
 z=g_\psi([\mu_x;v_x])\in\mathbb R^{16},
 \label{eq:conditioning}
\end{equation}
where $g_\psi$ is a learnable MLP with widths $8192\to512\to512\to16$ and SiLU activations. The MLP learns jointly with the healthy predictor and is then frozen. No personality labels are required.

For a predictor activation $h$, FiLM applies $\gamma(z)\odot h+\beta(z)$, initialized with gain one and bias zero. Here $s_{\le t}$ supplies the changing state history, while $z$ supplies context. Both condition the next-state prediction through shared model parameters. For offline screening, causal attention limits the state-history input, while $z$ summarizes the full recording.

To preserve the feature space used for healthy pretraining, IIP extracts states and conditioning statistics from the unadapted frozen backbone $\Phi$. NSP reads its decision state from the LoRA-adapted configuration $\Phi_\phi$. Thus the healthy reference retains its pretraining feature source during supervised NSP adaptation. The adapted NSP decision state is not reused as an IIP trajectory.

\label{sec:predictor}

We form the reference pool $\mathcal H=\{i:(\mathcal X_i,y_i)\in\mathcal D_{\mathrm{train}},\,y_i=0\}$ from the training set's non-depressed records. Here ``healthy'' refers to that dataset label. A shared predictor learns the next-state patterns within this pool. There is no per-subject control matching. Our world-model-inspired latent predictor \citep{ha2018worldmodels,lecun2022path,assran2025vjepa2} produces $\hat s_{t+1}=f_\theta(s_{\le t};z)$ from the observed history and context. Independently, the next observation is encoded as $\tilde s_{t+1}=E'(x_{t+1})$. We fit the predictor by
\begin{equation}
\begin{aligned}
\min_{\theta,E,\psi}\quad &\mathbb{E}_{\mathcal H}\big\lVert f_\theta(s_{\le t};z)-\sg(\tilde s_{t+1})\big\rVert_2^2\\
&+\lambda\,\mathcal{R}_{\mathrm{VICReg}}(E),
\end{aligned}
\label{eq:iip-objective}
\end{equation}
where $\sg$ blocks gradients through the target. The semi-orthogonal encoder $E'$ stays fixed, so prediction errors are measured against a consistent target space. The squared-error term learns to predict the independently encoded next observation. The variance term $\sum_j \max(0,1-\operatorname{std}(E_j))$ of $\mathcal{R}_{\mathrm{VICReg}}$ discourages low feature variance, while its covariance term reduces redundant coordinates \citep{bardes2022vicreg}.

For fixed representations and conditioning, squared-error learning targets the average next state of healthy training records with that history and context. The learned reference estimates this conditional pattern subject to data coverage and model capacity (\Cref{sec:theory}). It is a prediction along the same observed recording, not an observed healthy trajectory for that person.

A shared predictor accommodates different contexts without fitting a separate model to each person. It provides a common construction for comparing long recordings without segment-level emotion annotations.

The predictor uses three causal Transformer layers, expanded in \Cref{fig:framework}. Within the state-history input, causal attention restricts each position to its available prefix, while $z$ provides context conditioning. Each prediction is evaluated against the independently encoded next observation. At the following step, the predictor again receives the observed history, rather than recursively feeding back its previous prediction. Thus residual peaks localize one-step disagreement along the recording, without accumulating errors from a free-running rollout.

\subsection{Reference-derived emotional dynamics descriptors}
\label{sec:residual}
\label{sec:inertia}
We describe each recording through three complementary properties: departure from a healthy reference, persistence of residual fluctuations and the predictor's local spectral structure. These reference-derived emotional dynamics descriptors are recording-level proxies computed from latent behavioral states. Emotional dynamics supplies the theoretical motivation, while the measured quantities remain prediction errors and model properties.

\textbf{Reference Deviation ($R$)} measures average disagreement with the healthy predictor. For the observed and predicted next states,
\begin{equation}
\begin{aligned}
r_t&=\lVert\tilde s_{t+1}-\hat s_{t+1}\rVert_2,\\
R&=\frac{1}{T-1}\sum_{t=1}^{T-1}\lVert\tilde s_{t+1}-\hat s_{t+1}\rVert_2.
\end{aligned}
\label{eq:residual}
\end{equation}
A larger $R$ means that the observed continuation is less consistent with the reference. Averaging over transitions keeps the score from increasing simply because a recording is longer. Unlike within-person variability or successive-change measures, it compares observations with predictions, rather than with their own mean or adjacent observations. Predictive disagreement can also contain observation uncertainty, as formalized by the bias--noise decomposition in \Cref{sec:theory}. Its association with depression is evaluated by the supervised readout rather than assumed from the residual alone.

\textbf{Residual Decay Rate ($\forcear$)} characterizes the persistence of fluctuations in prediction disagreement. We center the residuals at their recording median, $e_t=r_t-\operatorname{median}_{\tau}r_\tau$, and fit a no-intercept AR(1) coefficient over adjacent valid pairs:
\begin{equation}
\hat\varphi=\frac{\sum_t e_te_{t+1}}{\sum_t e_t^2},\qquad
 a=-\ln\varphi_{\mathrm{clip}}.
\label{eq:residual-ar}
\end{equation}
We set $\varphi_{\mathrm{clip}}=\operatorname{clip}(\hat\varphi,0.01,0.999)$. The resulting decay rate is regularized toward the healthy-training prior $a_0$:
\begin{equation}
\forcear=\frac{a+2a_0}{3}.
\label{eq:ar-shrinkage}
\end{equation}
An unavailable fit receives $a_0$. Under a stationary AR(1) interpretation, $\varphi^k=\exp(-ka)$ describes decay over $k$ one-second steps. Thus \textbf{a smaller decay rate indicates stronger residual persistence}, or greater stickiness of predictive departure \citep{kuppens2010inertia,koval2013affect}. Unlike $R$, it concerns fluctuations around the recording's residual level. A large but rapidly varying residual need not be persistent. Because the fitted sequence is a residual magnitude, this is an inertia-inspired proxy for predictive-departure persistence, rather than a direct measure of the persistence of positive or negative affect.

\textbf{Jacobian Spectral Descriptor ($\rho(J)$)} summarizes the healthy predictor's local mapping in its learned coordinates. With mean online state $\bar s=T^{-1}\sum_{t=1}^{T}s_t$ and fixed context $z$, define the single-token map $F_z(s)=f_\theta([s];z)$. Its Jacobian and spectral radius are
\begin{equation}
J=\left.\frac{\partial F_z(s)}{\partial s}\right|_{s=\bar s},\qquad
\rho(J)=\max_j|\lambda_j(J)|.
\label{eq:jacobian}
\end{equation}
We estimate this property using 30 Jacobian--vector power iterations, avoiding an explicit $256\times256$ matrix. The estimate describes the reference predictor in its fixed online-to-target coordinates. \Cref{sec:theory} explains the distinction from a common-coordinate relaxation model.

\subsection{Personalized screening with two pathways}
\label{sec:nsp}
\label{sec:nsp-readout}
\label{sec:fusion}

NSP adapts the frozen omni-modal backbone with LoRA \citep{hu2022lora} and casts the risk judgment as the generation of a label token (\Cref{fig:modules}a). Let $h_{\mathrm{pos}}$ be the decision state, the hidden state one step before the label is written, and let $W\in\mathbb{R}^{d\times2}$ give the scores of the two risk tokens, represented by the label tokens \texttt{0} and \texttt{1}. The label $y\in\{0,1\}$ is supervised by cross-entropy,
\begin{equation}
\mathcal L_{\mathrm{CE}}=-\log p(\langle\mathrm{risk}_y\rangle\mid h_{\mathrm{pos}}),
\label{eq:ce}
\end{equation}
the adapter's only label-supervised path. Let $\ell_j=h_{\mathrm{pos}}^\top W_j$ be the logit of label token $j$. The population margin is $m=\ell_1-\ell_0=h_{\mathrm{pos}}^\top(W_1-W_0)$. Reading both scores at the same decision position also gives $m=\log(p_1/p_0)$, since their softmax normalizer cancels. Thus $m>0$ favors the positive label and $m=0$ marks equal label scores. ``Population'' refers to the shared readout parameters, not an average across subjects: each recording still produces its own margin. This continuous score retains the relative evidence needed for the additive correction. A thresholded label would discard it.

The population margin becomes personalized by merging it with the three IIP features (\Cref{fig:framework}, right). Writing $u=[R,\ \forcear,\ \rho(J)]$, we standardize it elementwise as $\tilde u=(u-\mu_u)\oslash\sigma_u$ using training-set statistics. The head converts the three standardized summaries into one correction. Adding this correction to the population score gives
\begin{equation}
b=w^\top\tilde u+c,\qquad \ell=m+b,\qquad \hat y=\mathbb 1[\ell\ge0].
\label{eq:decision}
\end{equation}
To learn the population score and its correction with separate gradient paths, we use
\begin{equation}
\mathcal L=\mathcal L_{\mathrm{CE}}+\mathrm{BCE}\big(\sigma(\sg(m)+b),\,y\big).
\label{eq:joint}
\end{equation}
The stop-gradient on $m$ leaves its value available to BCE while blocking its gradient to NSP. The decision can equivalently be written as $m\ge-b$: positive $b$ lowers the required screening margin, whereas negative $b$ raises it. Since $\sigma(m)=p_1/(p_0+p_1)$ is the relative two-label probability, adding $b$ multiplies the corresponding odds by $\exp(b)$. Here $c$ is a shared intercept, while $w^\top\tilde u$ supplies the recording-dependent adjustment. Two recordings with the same NSP margin can therefore receive different decisions when their reference features differ. The coefficient of $m$ is fixed at one: the head learns an additive adjustment, without separately learning how much to weight the population margin. The standardized vector $\tilde u\in\mathbb R^3$ is distinct from the predictor condition $z\in\mathbb R^{16}$. IIP alone predicts from $u$ without $m$, using its own supervised readout. \Cref{tab:ablation} examines the contribution of its feature subsets.

Standardization lets $w$ learn the contribution signs on a common training-defined scale. CE updates LoRA, while BCE learns $(w,c)$ from the current margin without changing the adapter's CE gradient. The modules $E$, $E^{\prime}$, $g_\psi$ and $f_\theta$ remain frozen.

The procedure therefore has three stages: healthy labels select the pool for self-supervised prediction, the frozen predictor supplies training-standardized features, and supervised learning fits the screening margin and its correction. Freezing predictor parameters still permits input differentiation for $J$. \Cref{alg:inertiampdd} gives the full training and inference procedure.

\begin{algorithm}[t]
\footnotesize
\caption{InertiaMPDD: training and inference}
\label{alg:inertiampdd}
\begin{algorithmic}[1]
\REQUIRE Training data $\mathcal D$, healthy pool $\mathcal H$, frozen backbone $\Phi$ and target encoder $E'$
\ENSURE IIP modules $(E,g_\psi,f_\theta)$, LoRA $\phi$, head $(w,c)$
\STATE \textbf{1. Learn the healthy reference}
\FOR{each healthy minibatch (600 updates)}
  \STATE Extract and align frozen features $x_{1:T}$ at $1$~Hz
  \STATE $s_t\leftarrow E(x_t)$, $\tilde s_t\leftarrow E'(x_t)$
  \STATE $z\leftarrow g_\psi([\mu_x;v_x])$, $\hat s_{t+1}\leftarrow f_\theta(s_{\le t};z)$
  \STATE Update $(E,g_\psi,f_\theta)$ by \Cref{eq:iip-objective}
\ENDFOR
\STATE \textbf{2. Extract descriptors with frozen IIP modules}
\STATE Estimate decay prior $a_0$ from healthy training records
\FOR{each training record}
  \STATE Recompute states, context and next-state predictions
  \STATE Compute $r_t$ and mean $R$ by \Cref{eq:residual}
  \STATE Fit AR(1) to $r_t-\operatorname{median}_{\tau}r_\tau$
  \STATE Clip $\varphi$ to $[0.01,0.999]$, set $a\leftarrow-\ln\varphi$
  \STATE $\forcear\leftarrow(a+2a_0)/3$ (or $a_0$ if unavailable)
  \STATE Estimate $\rho(J)$ at $\bar s$ with fixed $z$, 30 JVP steps
  \STATE $u\leftarrow[R,\forcear,\rho(J)]$
\ENDFOR
\STATE Fit training statistics $(\mu_u,\sigma_u)$ for $\tilde u$
\STATE \textbf{3. Learn screening and correction}
\FOR{each labeled minibatch}
  \STATE Read $m=\ell_1-\ell_0$ from adapted backbone $\Phi_\phi$
  \STATE $b\leftarrow w^\top\tilde u+c$
  \STATE Minimize \Cref{eq:joint}: CE updates $\phi$, BCE $(w,c)$
\ENDFOR
\STATE \textbf{Inference:} extract $u$ and $m$ using the learned modules
\STATE $\hat y\leftarrow\mathbb1[m+w^\top((u-\mu_u)\oslash\sigma_u)+c\ge0]$
\end{algorithmic}
\end{algorithm}

\input{sec/method_computations}

\input{sec/method_analysis}

\FloatBarrier

\section{Experiments}
\label{sec:experiments}

\subsection{Experimental setup}
\label{sec:setup}

D-Vlog \citep{yoon2022dvlog} and LMVD \citep{lmvd2025} contain spontaneous, self-recorded vlogs, evaluated here with binary depression labels. D-Vlog uses English YouTube recordings with the original test assignment restricted to currently available videos. LMVD uses Chinese vlogs and a $7{:}1{:}2$ split. The healthy predictor is trained only on healthy training recordings, with no uploader shared between training and evaluation.
AVEC2014 \citep{valstar2014avec} and MPDD \citep{fu2025first} provide interview/task-elicited recordings. Our AVEC2014 evaluation uses a subject-isolated $80{:}20$ split of 83 participants (67 training, 16 test), keeping all recordings and segments from each participant together. MPDD quantitative experiments use cohort-specific retraining with subject-isolated $80{:}20$ splits. Speech is transcribed with Whisper large-v3 \citep{whisper2023}. Checkpoints are selected on validation data.

Following the baseline evaluation protocol, we report accuracy and positive-class F1, with depression as the positive class. The pathway and feature results in \Cref{tab:main,tab:ablation} are multi-seed summaries over five seeds. Figures show single-run exports. Group differences and feature-performance significance are assessed by permutation tests across all datasets (\Cref{sec:statistics}). Published baselines on the two vlog corpora include Mamba-based sequence models, transformer-based fusion and DepressionAgent, with values from the original papers in the comparison format of \citet{depressionagent2026}. For AVEC2014/MPDD, baseline entries are selected by best results for each method and cohort.

Implementation settings for Qwen2.5-Omni-3B \citep{qwen25omni}, the predictor and the correction head are summarized in \Cref{tab:impl}.

\input{sec/implementation}

\input{sec/statistics}

\FloatBarrier

\subsection{Pathway complementarity across recording settings}
\label{sec:main}

\input{tables/main_results}

\textbf{Vlog gains indicate complementary reference evidence.} On D-Vlog, NSP reaches F1 $97.38$ and IIP alone $68.97$, while fusion reaches $98.41$ (\Cref{tab:main}). LMVD has more balanced pathways: IIP and NSP reach $75.88$ and $74.57$, and fusion reaches $84.98$, a $9.10$-point gain over its stronger pathway. Self-recorded vlogs permit individually chosen topics and spontaneous expression \citep{yoon2022dvlog,lmvd2025}. Direct population cues are already strong on D-Vlog, whereas LMVD leaves more scope for a context-conditioned healthy comparison. Thus standalone reference strength and complementarity with NSP are distinct sources of screening value.

\textbf{Interview settings can favor reference-derived descriptors.} IIP reaches F1 $92.31$ on MPDD-elder and $90.91$ on young, exceeding fusion at $85.71$ and $80.00$. Elder fusion still improves NSP by $8.79$ points. Guided interaction in MPDD \citep{fu2025first} can constrain freely expressed topics and behavior, making reference comparison informative alongside population classification. The ordering also motivates examining fusion separately: $m$ has coefficient one in \Cref{eq:decision}, so the correction head shifts the boundary without learning a separate pathway weight.

\textbf{AVEC2014 reveals a boundary of reference discrimination.} Fusion reaches $66.67$ F1 versus $50.00$ for either pathway, but remains below the largest displayed baseline value of $74.07$. Although AVEC2014 and MPDD both use elicited recordings, their pathway orderings differ. Task coverage, age, language and cohort composition therefore matter alongside acquisition setting. A useful healthy reference must cover the evaluated behavior. The group and control experiments below examine where its predictive departures carry discriminative information.

\subsubsection{Elicited recordings require task-matched references}
\label{sec:recording-settings}
MPDD uses guided interaction, while AVEC2014 includes prompted responses and reading \citep{fu2025first,valstar2014avec}. Reference coverage must therefore account for task regularities as well as expression, with age, language and cohort composition also varying across datasets.

\begin{table}[htbp]
\centering\small
\caption{Cross-split reference-deviation contrasts and interpretation. Differences are depressed minus healthy means.}
\label{tab:elicited-contrast}
\begin{tabular}{@{}lrr@{}}
\toprule
Cohort & Training $\Delta R$ & Test $\Delta R$ \\
\midrule
MPDD-elder & $1.571$ & $1.976$ \\
MPDD-young & $2.269$ & $2.554$ \\
AVEC2014 & $0.317$ & $-0.170$ \\
\bottomrule
\end{tabular}
\end{table}
The unstandardized contrasts retain the same direction across splits on MPDD, whereas AVEC2014 reverses direction. This links the screening differences to reference consistency and motivates broader task coverage.

\subsubsection{Paired intervals quantify decision-level uncertainty}
\input{tables/paired_uncertainty}
Paired bootstrap intervals quantify uncertainty in fixed-export gains. AVEC2014 and MPDD-young have wider intervals than D-Vlog, consistent with their smaller evaluation cohorts.

\subsection{From reference features to screening evidence}
\label{sec:groupdiff}
\label{sec:ablation}

\input{tables/group_differences}
\input{tables/feature_ablation}

\textbf{Vlog descriptors distinguish group shifts from individual discrimination.} D-Vlog and LMVD show higher depressed-group test residuals ($p=0.0003$ and $0.0040$), yet LMVD's residual-only screening is nonsignificant ($p=0.507$). Its Jacobian descriptor is the strongest single feature (F1 $69.47$), and the combination reaches $75.88$. D-Vlog's combination improves residual-only F1 from $56.77$ to $68.97$ despite weak marginal temporal and spectral contrasts. Group tests measure class-mean differences, whereas feature-performance tests assess prediction against shuffled labels. Joint descriptor information can therefore improve classification without significant shifts in every marginal mean.

\textbf{Reference deviation is the clearest recurring MPDD descriptor.} Depressed-group $R$ is higher in both training and test for both cohorts, with test $p=0.0005$ (elder) and $0.0004$ (young). Residual-only F1 reaches $82.35$ and $95.65$ (\Cref{tab:ablation}). Combining descriptors raises elder F1 to $92.31$ but gives $90.91$ on young. The recurring residual contrast supports reference departure as useful evidence, while the value of temporal and spectral additions depends on the cohort.

\textbf{AVEC2014 lacks a stable cross-split signature.} The depressed-minus-healthy residual contrast changes from $+0.317$ in training to $-0.170$ in test, with neither significant. The test-only spectral contrast ($p=0.0265$) does not recur significantly in training, and spectral-only screening remains nonsignificant ($p=0.5110$). Combined F1 is $50.00$ ($p=0.2846$), locating the main difficulty in reference discrimination rather than simply pathway combination.

\textbf{Duration can interact with the behavioral descriptors.} Adding it raises D-Vlog F1 to $72.00$ but reduces accuracy, and gives no F1 gain on LMVD. Duration-augmented MPDD readouts reach $93.33$ and $100.00$ F1 despite weak duration-only performance. We keep duration outside the main IIP definition. Duration-matched and healthy-pool controls further examine these interactions (\Cref{tab:reference-pool}).

\subsubsection{Candidate proxy markers and their scope}
\label{sec:marker-scope}
\label{sec:discussion}
\label{sec:division}
\label{sec:dynamics}
The results support reference-derived emotional dynamics as \emph{candidate proxy biomarkers}: computable behavioral descriptors associated with depression labels under recording-level supervision. Their relevance is established through screening, rather than direct emotion annotations. Stress, fatigue and acquisition conditions may also affect predictive departure. Comparison with alternative conditions can test specificity, while repeated recordings and longitudinal symptom measurements can assess repeatability and sensitivity to change within a defined context of use \citep{fdabiomarkers}. The cohort-specific fits evaluate within-cohort utility, while a fixed reference on an unseen cohort requires a distinct test of transfer.

\input{sec/feature_distributions}

\subsection{Component ablation}
\label{sec:design-controls}
\input{tables/design_controls}

\Cref{tab:design-controls} summarizes decision and reference-construction ablations. The comparisons below separate the effects of context conditioning, temporal order, modality and pathway combination.

\input{sec/experimental_controls}

\input{sec/numerical_validation}

\subsection{Case studies and visualization}
\label{sec:case}
\label{sec:score-geometry}

\begin{figure}[t]
  \centering
  \includegraphics[width=\linewidth]{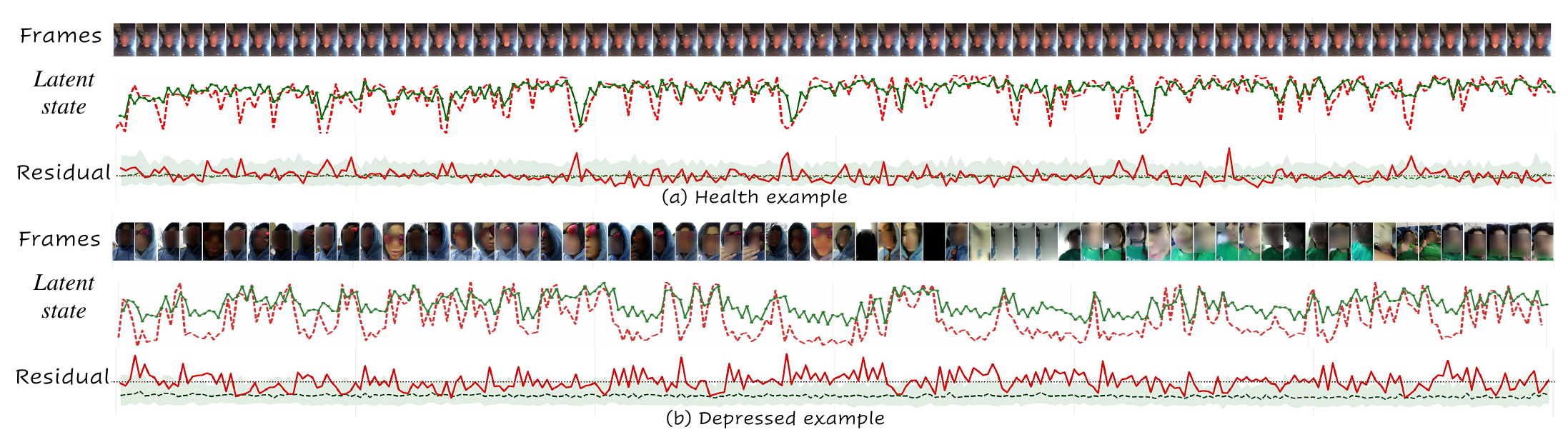}
  \caption{\textbf{D-Vlog case studies.} Healthy (top) and depressed (bottom) recordings. Green: healthy reference. Red: subject trace or residual. Shading: healthy residual 5--95\% range.}
  \label{fig:dvlog-visualization}
\end{figure}

\begin{figure}[!t]
  \centering
  \includegraphics[width=0.90\linewidth]{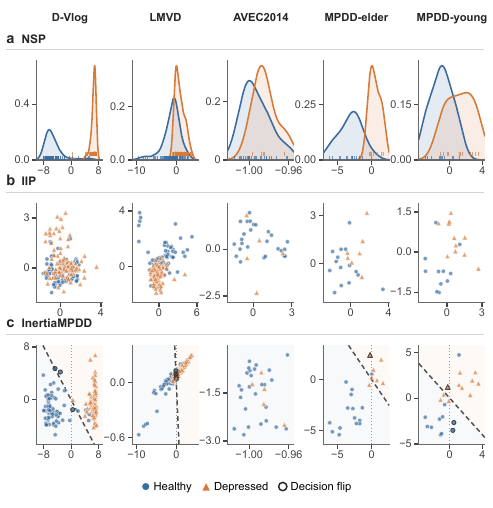}
  \caption{\textbf{Test-set pathway views from single-run exports.} Columns: cohorts. Rows: NSP margins, IIP features and InertiaMPDD $(m,b)$ with boundary $b=-m$. Blue circles: healthy. Orange triangles: depressed. Dark outlines: changed decisions.}
  \label{fig:pathway-scatter}
\end{figure}

\textbf{Residual traces localize predictive disagreement.} \Cref{fig:dvlog-visualization} illustrates the temporal information compressed into $R$. The healthy trajectory follows its reference through marked fluctuations, whereas the depressed example shows sustained residual elevations. Movement alone is insufficient: a changing trajectory can be well predicted, while a smooth one can remain offset from its reference. The full-state residual localizes disagreement even when the displayed projection looks similar. Recording-level labels identify the screening category, while residual traces indicate when the observed continuation differs from the learned reference. These traces guide audiovisual inspection of isolated and recurring disagreement. Averaging these errors yields $R$, and residual decay summarizes the persistence of their fluctuations. Together with the spectral descriptor, these summaries support recording-level screening. They locate predictive disagreement without segment-level symptom attribution. \Cref{sec:multi-dataset-cases} includes the complete multi-dataset cases.

\textbf{Correction depends on joint score geometry.} \Cref{fig:pathway-scatter} connects the descriptors to screening decisions. The top row shows continuous NSP margins. The bottom row places each recording relative to $b=-m$. D-Vlog separates strongly along $m$, while its IIP view overlaps substantially. LMVD leaves more scope for adjustment in the plotted run. \Cref{tab:main} reports the quantitative pathway comparison. In the plotted AVEC2014 run, both classes lie on the negative-score side, so feature variation alone does not ensure a useful decision boundary. On MPDD-elder, correction moves one missed positive across the boundary. It improves NSP without surpassing standalone IIP. Marginal densities describe coordinate separation, whereas the joint $(m,b)$ location determines the additive decision.

For a schematic example, recordings with the same $m=0.4$ but corrections $b=-0.7$ and $b=0.2$ receive final scores $-0.3$ and $0.6$. The recording-dependent correction therefore changes the decision even when population evidence is identical.

\input{sec/multi_dataset_cases}

\FloatBarrier

\section{Conclusion}
\label{sec:conclusion}

InertiaMPDD explores how pretrained multimodal representations can supply measurable evidence for depression screening without fine-grained emotion labels. Its world-model-inspired healthy reference produces reference-derived emotional dynamics descriptors for standalone screening and individual correction. The experiments support three findings. \textbf{A predictive reference makes behavioral departure useful for screening}, with recurring MPDD deviation contrasts and complementary vlog evidence. \textbf{Reference construction draws on both temporal order and multimodal content}: LMVD benefits from ordered prediction and joint audio--visual inputs, while strong static controls on MPDD-young show that informative reference evidence can also reside in recording-level state patterns. \textbf{Reference informativeness and fusion benefit remain distinct}, motivating cohort-sensitive use of the two pathways. AVEC2014's weak standalone separation identifies reference coverage as a key boundary. These candidate proxy descriptors warrant further validation of their specificity, repeatability and transfer.

%% file: sec/related_work.tex
\section{Related Work}
\label{sec:related}

\subsection{A predictive reference complements population screening}
Depression screening has progressed from hand-crafted paralinguistic features to multimodal sequence models. DAIC-WOZ and AVEC established interview-based benchmarks \citep{gratch2014daic,valstar2014avec}. Recent methods model temporal dependencies, cross-modal interactions and individual context. Examples include MMFformer \citep{mmfformer2025}, CAF-Mamba \citep{cafmamba2026}, DynMultiDep \citep{dynmultidep2026}, ASYM \citep{asym2026}, AttMamba \citep{attmamba2026} and personality-aware PAVNet \citep{pavnet2025}. DepressionAgent provides symptom-level explanations through multimodal reasoning \citep{depressionagent2026}. IIP adds a predictive reference against which the observed continuation can be compared. This supplies a time-resolved departure signal alongside the population decision.

Personality-aware benchmark design includes the MPDD challenge \citep{fu2025first} and MPDD-AVG, which combines audio-visual interviews with gait analysis \citep{fu2026mpdd}. Physiological measurements offer a complementary route: M$^3$ADD organizes multimodal, multitask and multievent depression assessment \citep{fu2025m}, while M3DEP studies individual differences in EEG--EMG signals \citep{fu2026m3dep}. These studies broaden the modalities and individual factors considered in screening. IIP examines recording-dependent variation through a healthy predictive reference in pretrained behavioral representations.

Prediction-based anomaly detection also learns regular behavior and measures future-prediction error \citep{liu2018prediction}. Our question is whether a healthy-trained reference, conditioned on context, yields useful descriptors for depression screening. Counterfactual approaches address related questions about spurious associations, such as gender in acoustic prediction \citep{genderdebias2026}. Here the comparison is operationally between a predicted next target state and the next observation in the same recording.

\subsection{World-model-inspired prediction supplies a learnable reference}
World models learn representations and predict how observations evolve \citep{ha2018worldmodels}. Joint-embedding predictive architectures apply prediction in representation space \citep{lecun2022path,assran2023ijepa}. V-JEPA 2 combines action-free predictive pretraining with action-conditioned learning for planning \citep{assran2025vjepa2}. IIP adapts latent prediction to a healthy behavioral reference for screening, with variance and covariance regularization \citep{bardes2022vicreg}. The evaluation concerns reference-derived descriptors rather than free-running simulation or planning.

\subsection{Emotional inertia motivates residual-persistence analysis}
Emotional inertia concerns the persistence of affect from one moment to the next, commonly described by lagged autocorrelation. It has been associated with depression \citep{kuppens2010inertia,koval2013affect}, and meta-analytic work connects affective dynamics with psychological well-being \citep{houben2015meta}. Critical slowing down provides a related motivation for examining persistence and recovery around depressive transitions \citep{vandeleemput2014critical}. These findings motivate our residual AR analysis. Its measured sequence is prediction disagreement in learned audiovisual coordinates. We therefore distinguish the affective construct from its computational proxy: the AR analysis describes persistence in prediction disagreement, without assigning that disagreement an emotional valence.

Recovery and reactivity additionally require an identified baseline or eliciting event, neither of which is directly measured here. Context conditioning also does not by itself identify stable psychological traits separately from momentary states \citep{hyvarinen2019nonlinear,scholkopf2021causal}.

%% file: sec/method_computations.tex
\subsection{Detailed NSP and IIP Computations}
\label{sec:computations}

\begin{figure}[!t]
\centering
\includegraphics[width=\textwidth]{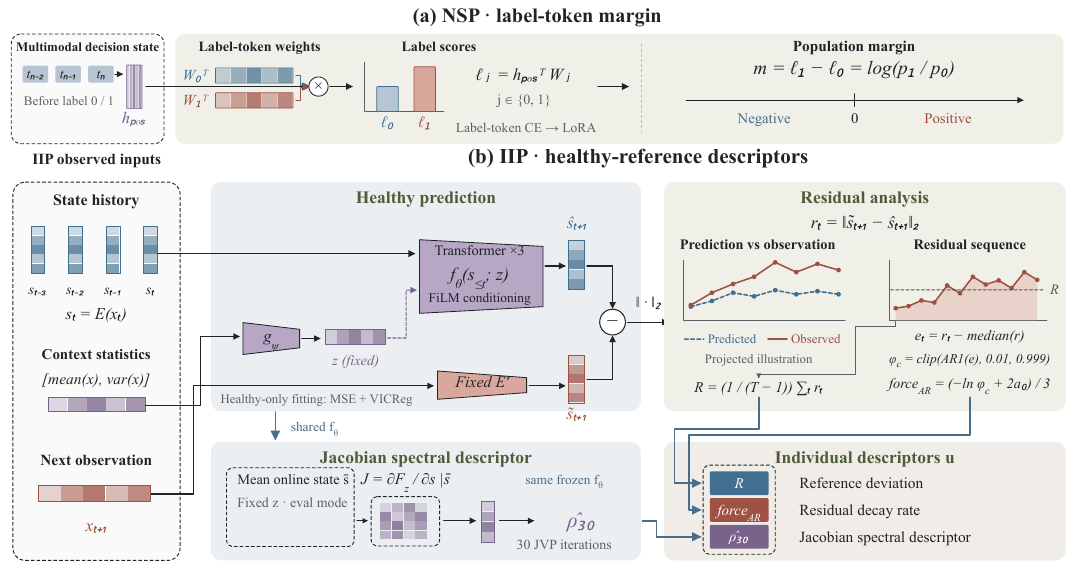}
\caption{\textbf{Detailed pathway computations.} (a) NSP label-score readout. (b) IIP healthy prediction, residual analysis, Jacobian spectral estimation and descriptor assembly. Tokens, curves and matrix cells are schematic.}
\label{fig:modules}
\end{figure}

\Cref{fig:modules} expands the two computational routes. In NSP, the highlighted weights $W_0$ and $W_1$ read label scores from the multimodal state immediately before the output digit. In IIP, history and context feed the healthy predictor, while the next observation follows the independent fixed-target route. The blue dashed and red solid curves show a projection of the prediction and observation. Their residual uses every target-space coordinate.

The residual-analysis route supplies $R$ and $\forcear$. The Jacobian route instead differentiates the frozen predictor at $\bar s$ with $z$ fixed, supplying $\hat\rho_{30}$. These routes meet at the descriptor vector $u$, whose training-standardized values enter the correction head. Thus $z$ conditions the reference, while $u$ summarizes its predictive comparison and local mapping. The spectral route remains separate from the observed residual sequence.

%% file: sec/method_analysis.tex
\subsection{Analytical properties of the reference}
\label{sec:theory}

Let $s_t=E(x_t)\in\mathbb R^{256}$ be the online state, $\tilde s_t=E'(x_t)$ the fixed-target state, and $z\in\mathbb R^{16}$ the context vector. The following results explain which mathematical properties motivate the descriptors and how the decision gradients are separated.

\subsubsection{Residuals combine predictive departure and observation noise}
\begin{proposition}[Bias--noise decomposition of the residual]
\label{prop:residual}
Let $\tilde s_{t+1}=g(s_{\le t};z)+\varepsilon_t$, with $\mathbb{E}[\varepsilon_t\mid s_{\le t},z]=0$ and conditional covariance $\Sigma_\varepsilon(s_{\le t},z)$. Then
\begin{equation}
\begin{aligned}
\mathbb{E}[r_t^2\mid s_{\le t},z]
&=\underbrace{\lVert g(s_{\le t};z)-f_\theta(s_{\le t};z)\rVert_2^2}_{\text{departure term } \Delta_t^2}\\
&\quad+\operatorname{tr}(\Sigma_\varepsilon(s_{\le t},z)).
\end{aligned}
\label{eq:decomposition}
\end{equation}
When $g\approx f_\theta$, the expected squared residual approaches the conditional noise trace. A larger departure term increases the expected squared residual at a fixed noise level.
\end{proposition}

\paragraph{Proof of \Cref{prop:residual} (bias--noise decomposition).}
Substituting the observation model and suppressing the shared function arguments gives
\begin{align}
\mathbb{E}\big[r_t^2\mid s_{\le t},z\big]
&= \mathbb{E}\big[\lVert g+\varepsilon_t-f_\theta\rVert_2^2 \mid s_{\le t},z\big] \nonumber\\
&= \lVert g-f_\theta\rVert_2^2
 + 2\,\mathbb{E}[\varepsilon_t\mid s_{\le t},z]^\top\big(g-f_\theta\big)
\nonumber\\
&\quad+ \mathbb{E}\big[\lVert\varepsilon_t\rVert_2^2\mid s_{\le t},z\big] \nonumber\\
&= \lVert g-f_\theta\rVert_2^2 + \operatorname{tr}(\Sigma_\varepsilon(s_{\le t},z)),
\end{align}
where $\mathbb E[\varepsilon_t\mid s_{\le t},z]=0$ eliminates the cross term. The recording-level feature $R$ averages the corresponding unsquared norms. \hfill$\square$

The decomposition separates mismatch with the learned healthy reference from observation uncertainty. The downstream screening task evaluates the relevance of their combined residual.

\subsubsection{AR persistence motivates the regularized decay descriptor}
\begin{proposition}[Residual AR(1) decay]
\label{prop:ar1}
Let the stationary, mean-centered sequence $e_t=r_t-\mathbb E[r_t]$ follow $e_{t+1}=\varphi e_t+\eta_t$ with innovations satisfying $\mathbb E[\eta_t\mid e_t,e_{t-1},\ldots]=0$ and $0<\varphi<1$. Its lag-one autocorrelation is $\varphi$, and $\mathbb E[e_{t+k}\mid e_t]=\varphi^k e_t=\exp(-ka)e_t$, where $a=-\ln\varphi$ is the unregularized residual decay rate.
\end{proposition}

This stationary, mean-centered model motivates the decay interpretation. The implemented median-centering, clipping and shrinkage regularize the empirical descriptor without establishing stationarity.

\paragraph{Proof of \Cref{prop:ar1}.}
Under the stationary AR(1) and conditional-mean assumptions, $\operatorname{Cov}(e_{t+1},e_t)=\varphi\operatorname{Var}(e_t)$, so its lag-one autocorrelation is $\varphi$. Repeated conditional expectation gives $\mathbb E[e_{t+k}\mid e_t]=\varphi^k e_t$. Substituting $a=-\ln\varphi$ yields the exponential form. \Cref{eq:ar-shrinkage} then regularizes this estimate toward the healthy-training prior before it enters the screening features. \hfill$\square$

\subsubsection{The Jacobian spectral descriptor depends on the learned coordinates}
\begin{proposition}[Common-coordinate transition: conditional relaxation result]
\label{prop:relaxation}
For a first-order transition in common state coordinates, let $\bar s$ be a fixed point and $\delta_t=s_t-\bar s$. To first order $\delta_{t+k}\approx J^k\delta_t$. For diagonalizable $J$ with $0<\rho(J)<1$, its slowest linearized decay has characteristic time $\tau=-1/\ln\rho(J)$, which diverges as $\rho(J)\to1^-$.
\end{proposition}

\paragraph{Proof of \Cref{prop:relaxation}.}
Under the common-coordinate first-order transition assumptions of \Cref{prop:relaxation}, a Taylor expansion about the fixed point $\bar s$ gives $\delta_{t+1}\approx J\delta_t$, hence $\delta_{t+k}\approx J^k\delta_t$. For diagonalizable $J=V\Lambda V^{-1}$ we have $\lVert J^k\rVert\le\kappa(V)\,\rho(J)^k$ with $\kappa(V)$ the condition number of $V$, which bounds the linearized decay when $0<\rho(J)<1$. The slowest mode attains this rate. Solving $\rho(J)^{\tau}=e^{-1}$ gives $\tau=-1/\ln\rho(J)$. \hfill$\square$

The transition result assumes a common state basis and a fixed point. Our local Jacobian analysis differentiates the online-to-target map $F_z$ at the mean online state. Under independent coordinate changes $A$ and $B$, its Jacobian becomes $BJA^{-1}$, so eigenvalues can change. We therefore interpret the spectral estimate in the fixed learned coordinates. The mean state need not be an equilibrium, and non-normal transitions can show transient amplification even when their spectral radius is below one.

For IIP, $\rho(J)$ is therefore reported in the fixed learned coordinates rather than converted to a recovery time. \Cref{tab:spectral-validation} evaluates its numerical approximation.

\subsubsection{Stop-gradient separates adapter and correction learning}
\begin{proposition}[Role of the stop-gradient on the margin]
\label{prop:sg}
Assume $b$ is computed from the fixed unadapted IIP and is independent of the LoRA parameters. Without a stop-gradient on $m$, BCE contributes an adapter gradient. With $\sg(m)$, the adapter gradient equals the CE component at the current parameters.
\end{proposition}

\paragraph{Proof of \Cref{prop:sg}.}
Writing $\ell_{\mathrm{BCE}}=-y\log\sigma(m+b)-(1-y)\log(1-\sigma(m+b))$, we have $\partial\ell_{\mathrm{BCE}}/\partial m=\sigma(m+b)-y$. Without a stop-gradient this term reaches the LoRA parameters $\phi$ through $m$. With $\sg(m)$, BCE reaches only the bias-head parameters, so $\partial\mathcal L/\partial\phi=\partial\mathcal L_{\mathrm{CE}}/\partial\phi$ exactly. This identity isolates the two gradients while retaining the dependence of the bias-head update on the screening margin. \hfill$\square$

%% file: sec/implementation.tex
\subsection{Implementation and Evaluation Settings}
\label{sec:implementation}

\begin{table}[!t]
\centering
\small
\caption{Implementation and hyperparameters.}
\label{tab:impl}
\resizebox{\linewidth}{!}{%
\begin{tabular}{@{}ll@{}}
\toprule
Item & Value \\
\midrule
Backbone & Qwen2.5-Omni-3B thinker (bf16, frozen) \\
LoRA & rank 8, trained from scratch, lr $1\times10^{-4}$, batch 1 $\times$ grad-accum 8, linear schedule \\
Bias head & Linear($3\to1$), fp32, AdamW lr $1\times10^{-2}$ (separate parameter group), $\mathcal{N}(0,0.1)$ init \\
IIP & online encoder $4096\to256$, 3-layer FiLM causal Transformer, width 256, 4 heads, \\
    & feed-forward width 512, dropout 0.1, pre-normalization, fixed semi-orthogonal target \\
Conditioning MLP & $8192\to512\to512\to16$, SiLU, final-layer $\mathcal N(0,0.01)$ initialization \\
IIP optimization & Adam, lr $3\times10^{-4}$, batch 4, 600 steps, variance/covariance weights 0.1 each \\
Losses & NSP: CE, bias head: BCE (stop-gradient on the margin), IIP: MSE $+$ VICReg \\
Input limits & audio $\le300$\,s (25 tok/s), IIP visual sampling: 64 frames \\
Latent state & 1\,Hz pooled concatenation of audio 2048 and visual 2048 ($x_t\in\mathbb{R}^{4096}$) $\to s_t\in\mathbb{R}^{256}$ \\
\bottomrule
\end{tabular}}
\end{table}

\subsubsection{Temporal alignment and context conditioning}
Audio is limited to 300 seconds and provides 25 hidden tokens per second before one-second pooling. Visual alignment follows the grid defined in \Cref{sec:latent}.

The D-Vlog dynamics pool uses recordings of at least 120 seconds. This length criterion applies to reference learning, while screening evaluation retains the full test cohort.

\subsubsection{Residual decay and Jacobian spectral estimation}
The AR coefficient is fitted without an intercept after subtracting the recording median from the scalar residual sequence. Fewer than ten residuals or a lag-energy sum below $10^{-6}$ yields an unavailable estimate. Otherwise, the coefficient is clipped to $[0.01,0.999]$. Its negative logarithm lies between approximately $0.001$ and $4.605$ per one-second step. The prior $a_0$ is the median available raw decay in the healthy training group, with weight two in the shrinkage formula. An unavailable estimate receives $a_0$. The training prior is also used on held-out recordings.

For the spectral feature, the predictor is in evaluation mode, with dropout disabled. The input is the single-token state $[\bar s]$ and the condition $z$ is fixed. Autograd computes Jacobian--vector products without storing a dense Jacobian. Power iteration starts from a random vector with seed zero. An iterate norm below $10^{-12}$ returns zero. The updates and final estimate are
\begin{equation}
 v_{k+1}=\frac{Jv_k}{\lVert Jv_k\rVert_2},\qquad
 \hat\rho_{30}=\lVert Jv_{29}\rVert_2.
 \label{eq:power-iteration}
\end{equation}
The forward pass uses bf16. The reported descriptor is a finite-iteration spectral-radius estimate at the mean online state.

Standalone IIP fits its own supervised binary readout on $u$, without the NSP margin, with validation-based model selection.

%% file: sec/statistics.tex
\subsection{Feature Aggregation and Statistical Tests}
\label{sec:statistics}

\subsubsection{Group means summarize recording-level descriptors}
Each recording $i$ supplies three scalars: time-averaged residual $R_i$, shrinkage-adjusted residual decay $\forcear^{(i)}$, and the power-iteration estimate of $\rho(J_i)$ at its mean state. Within class $g$ and split $\mathcal S$, \Cref{tab:group} reports
\begin{equation}
\bar q_{g,\mathcal S}=\frac{1}{N_{g,\mathcal S}}\sum_{i\in\mathcal I_{g,\mathcal S}}q_i,
\label{eq:group-mean}
\end{equation}
where $q_i$ is one of these extracted, unstandardized features and $N_{g,\mathcal S}=|\mathcal I_{g,\mathcal S}|$. Each recording contributes one value, irrespective of its duration. Group means use unstandardized descriptors, while decision readouts standardize each recording separately using training statistics.

\subsubsection{Group and prediction tests address different hypotheses}
The tests in \Cref{tab:group} assess healthy--depressed differences in descriptor values. The tests in \Cref{tab:ablation} assess predictive performance against label permutations. A descriptor can therefore have a group mean contrast but limited individual discrimination. The vlog residual tests use the directional hypothesis of higher values in the depressed group.

%% file: tables/main_results.tex
\begin{table}[t]
\centering\footnotesize
\setlength{\tabcolsep}{2.5pt}
\caption{Depression-screening Accuracy and F1 score (\%). Bold: largest displayed value. --: unavailable.}
\label{tab:main}
\begin{adjustbox}{max width=\linewidth}
\begin{tabular}{@{\extracolsep{\fill}}lrrrrrrrrrr@{}}
\toprule
 & \multicolumn{2}{c}{D-Vlog} & \multicolumn{2}{c}{LMVD} & \multicolumn{2}{c}{AVEC2014} & \multicolumn{2}{c}{MPDD-elder} & \multicolumn{2}{c}{MPDD-young} \\
\cmidrule(lr){2-3}\cmidrule(lr){4-5}\cmidrule(lr){6-7}\cmidrule(lr){8-9}\cmidrule(lr){10-11}
Method & Accuracy & F1 score & Accuracy & F1 score & Accuracy & F1 score & Accuracy & F1 score & Accuracy & F1 score \\
\midrule
Depression Detector \citep{yoon2022dvlog} & -- & 63.50 & -- & -- & 72.00 & \textbf{74.07} & 64.90 & 56.90 & 59.09 & 60.87 \\
TAMFN \citep{tamfn2022} & -- & 65.82 & -- & -- & 64.00 & 70.97 & 65.22 & 20.00 & 59.09 & 60.87 \\
CAIINET \citep{caiinet2023} & -- & 66.55 & -- & -- & 60.00 & 53.80 & 45.50 & 50.00 & 81.10 & 82.40 \\
Depressformer \citep{depressformer2024} & 65.00 & 59.00 & -- & -- & \textbf{84.00} & 33.33 & 55.20 & 45.70 & 68.18 & 74.07 \\
DepMamba \citep{depmamba2025} & 68.87 & 76.44 & 72.13 & 73.20 & 80.00 & 16.67 & 78.26 & 61.54 & 77.27 & 80.00 \\
CMF-Mamba \citep{cmfmamba2025} & -- & -- & 74.86 & 75.53 & 46.70 & 27.30 & 52.30 & 44.80 & 74.10 & 76.00 \\
STCM-Mamba \citep{stcmmamba2025} & 72.64 & 77.69 & 75.41 & 76.49 & 62.00 & 68.85 & 60.87 & 30.77 & 63.64 & 63.64 \\
MSI-Net \citep{msinet2026} & 70.25 & 73.91 & 74.34 & 77.06 & 69.05 & 38.10 & 47.83 & 53.85 & 72.73 & 76.92 \\
MDDFormer \citep{lmvd2025} & -- & -- & 76.88 & 76.85 & 74.00 & 33.30 & 73.91 & 66.67 & 77.27 & 73.68 \\
BHDD \citep{bhdd2026} & -- & 75.94 & 78.11 & 78.06 & 62.00 & 67.80 & 66.36 & 17.78 & 52.27 & 51.16 \\
DepressionAgent \citep{depressionagent2026} & 95.26 & 95.69 & 82.57 & 83.04 & 56.25 & 58.82 & 78.26 & 44.44 & 59.09 & 30.77 \\
\midrule
NSP (alone) & 97.14 & 97.38 & 74.30 & 74.57 & 75.00 & 50.00 & 86.36 & 76.92 & 50.00 & 66.67 \\
IIP (alone) & 64.00 & 68.97 & 73.70 & 75.88 & 62.50 & 50.00 & \textbf{95.45} & \textbf{92.31} & \textbf{90.91} & \textbf{90.91} \\
\rowcolor{methodrow}
\textbf{InertiaMPDD} & \textbf{98.29} & \textbf{98.41} & \textbf{84.00} & \textbf{84.98} & 81.25 & 66.67 & 90.91 & 85.71 & 81.82 & 80.00 \\
\bottomrule
\end{tabular}
\end{adjustbox}
\end{table}

%% file: tables/paired_uncertainty.tex
\begin{table}[!htbp]
\centering\footnotesize
\setlength{\tabcolsep}{1.0pt}
\caption{Paired-bootstrap F1 intervals (\%). Differences are percentage points.}
\label{tab:paired-uncertainty}
\begin{tabular}{@{}llrr@{}}
\toprule
\rowcolor{tableheader}
Cohort & Quantity & F1 / $\Delta$F1 & 95\% CI \\
\midrule
D-Vlog & NSP & 97.38 & [94.84, 99.47] \\
 & Full & 98.41 & [96.38, 100.00] \\
 & Gain & +1.03 & [0.00, +2.76] \\
\midrule
AVEC2014 & NSP & 50.00 & [0.00, 85.71] \\
 & Full & 66.67 & [0.00, 100.00] \\
 & Gain & +16.67 & [$-50.00$, +80.00] \\
\midrule
MPDD-young & NSP & 66.67 & [42.86, 81.08] \\
 & Full & 80.00 & [54.52, 95.65] \\
 & Gain & +13.33 & [$-11.52$, +37.44] \\
\bottomrule
\end{tabular}
\end{table}

%% file: tables/group_differences.tex
\begin{table}[t]
\centering
\small
\renewcommand{\arraystretch}{1.08}
\setlength{\tabcolsep}{2.5pt}
\caption{Healthy and depressed group means of unstandardized IIP descriptors. Bold: $p<0.05$.}
\label{tab:group}
\begin{adjustbox}{max width=\linewidth}
\begin{tabular}{@{\extracolsep{\fill}}llrrrrrrrrr@{}}
\toprule
& & \multicolumn{3}{c}{Reference deviation $R$} & \multicolumn{3}{c}{Residual decay rate $\forcear$} & \multicolumn{3}{c}{\shortstack{Jacobian spectral\\descriptor $\rho(J)$}} \\
\cmidrule(lr){3-5}\cmidrule(lr){6-8}\cmidrule(lr){9-11}
Dataset & Split & Healthy & Depressed & $p$-value & Healthy & Depressed & $p$-value & Healthy & Depressed & $p$-value \\
\midrule
D-Vlog & Test & 6.707 & 6.901 & \textbf{0.0003} & 2.103 & 1.964 & 0.998 & 0.716 & 0.710 & 0.467 \\
 & Training & 6.820 & 6.916 & 0.078 & 2.090 & 1.998 & 0.897 & 0.718 & 0.709 & 0.421 \\
LMVD & Test & 12.510 & 12.887 & \textbf{0.0040} & 1.559 & 1.409 & \textbf{0.0002} & 0.938 & 0.731 & \textbf{$<$0.0001} \\
 & Training & 12.487 & 12.966 & \textbf{$<$0.0001} & 1.547 & 1.435 & \textbf{0.0010} & 0.898 & 0.750 & \textbf{$<$0.0001} \\
AVEC2014 & Test & 11.805 & 11.635 & 0.8142 & 1.634 & 1.779 & 0.3405 & 0.994 & 0.869 & \textbf{0.0265} \\
 & Training & 10.759 & 11.076 & 0.3830 & 2.004 & 1.969 & 0.5502 & 1.364 & 1.301 & 0.2868 \\
MPDD-elder & Test & 7.076 & 9.052 & \textbf{0.0005} & 0.466 & 0.469 & 0.8551 & 1.396 & 1.499 & 0.3009 \\
 & Training & 7.341 & 8.912 & \textbf{$<$0.0001} & 0.467 & 0.479 & 0.3829 & 1.400 & 1.367 & 0.4118 \\
MPDD-young & Test & 8.238 & 10.792 & \textbf{0.0004} & 0.683 & 0.679 & 0.9047 & 0.954 & 1.091 & 0.1359 \\
 & Training & 8.313 & 10.582 & \textbf{$<$0.0001} & 0.728 & 0.678 & \textbf{0.0119} & 1.038 & 1.068 & 0.5215 \\
\bottomrule
\end{tabular}
\end{adjustbox}
\end{table}

%% file: tables/feature_ablation.tex
\begin{table}[t]
\centering\footnotesize
\setlength{\tabcolsep}{1.2pt}
\caption{Feature ablations and duration controls (\%). Combined: $R+\forcear+\rho(J)$. Bold $p$-values: $p<0.05$.}
\label{tab:ablation}
\begin{adjustbox}{max width=\linewidth}
\begin{tabular}{@{\extracolsep{\fill}}lrrrrrrrrrrrrrrr@{}}
\toprule
 & \multicolumn{3}{c}{D-Vlog} & \multicolumn{3}{c}{LMVD} & \multicolumn{3}{c}{AVEC2014} & \multicolumn{3}{c}{MPDD-elder} & \multicolumn{3}{c}{MPDD-young} \\
\cmidrule(lr){2-4}\cmidrule(lr){5-7}\cmidrule(lr){8-10}\cmidrule(lr){11-13}\cmidrule(lr){14-16}
Features & Accuracy & F1 score & $p$-value & Accuracy & F1 score & $p$-value & Accuracy & F1 score & $p$-value & Accuracy & F1 score & $p$-value & Accuracy & F1 score & $p$-value \\
\midrule
\shortstack[l]{Reference deviation\\$R$} & 60.36 & 56.77 & \textbf{0.0015} & 57.80 & 59.69 & 0.507 & 25.00 & 33.33 & 0.9500 & 86.36 & 82.35 & \textbf{0.0005} & 95.45 & 95.65 & \textbf{0.0001} \\
\shortstack[l]{Residual decay rate\\$\forcear$} & 54.44 & 49.67 & 0.092 & 54.40 & 57.93 & \textbf{0.023} & 62.50 & 50.00 & 0.2846 & 59.09 & 47.06 & 0.3881 & 40.91 & 43.48 & 0.8969 \\
\shortstack[l]{Jacobian spectral\\descriptor $\rho(J)$} & 46.75 & 50.55 & 0.877 & 68.30 & 69.47 & \textbf{$<$0.0005} & 50.00 & 42.86 & 0.5110 & 68.18 & 58.82 & 0.1142 & 59.09 & 57.14 & 0.3352 \\
\rowcolor{methodrow}
Combined & 64.00 & 68.97 & \textbf{0.0005} & 73.70 & 75.88 & \textbf{$<$0.0005} & 62.50 & 50.00 & 0.2846 & 95.45 & 92.31 & \textbf{0.0002} & 90.91 & 90.91 & \textbf{$<$0.0001} \\
\midrule
\shortstack[l]{Combined\\$+$ duration} & 62.72 & 72.00 & \textbf{0.001} & 74.70 & 75.18 & \textbf{$<$0.0005} & 62.50 & 50.00 & 0.2846 & 95.45 & 93.33 & \textbf{0.0001} & 100.00 & 100.00 & \textbf{$<$0.0001} \\
Duration & 56.21 & 68.64 & 0.148 & 53.50 & 52.32 & 0.559 & 50.00 & 33.33 & 0.7154 & 68.18 & 46.15 & 0.2646 & 63.64 & 69.23 & 0.1835 \\
\bottomrule
\end{tabular}
\end{adjustbox}
\end{table}

%% file: sec/feature_distributions.tex
\subsection{Feature Distributions and Screening Implications}
\label{sec:feature-distributions}

\begin{figure}[!tp]
\centering
\includegraphics[width=\textwidth,height=0.86\textheight,keepaspectratio]{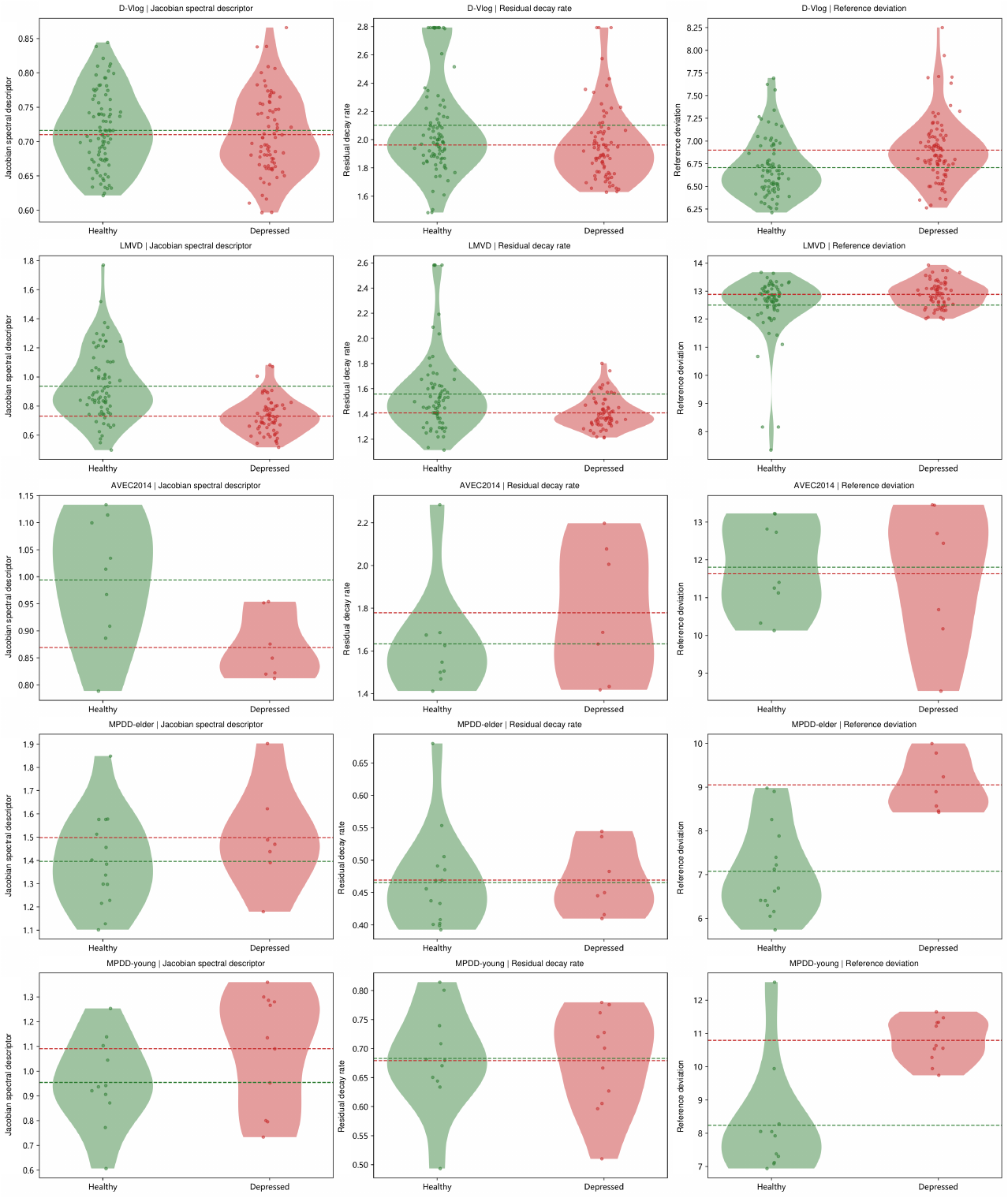}
\caption{\textbf{Test-set IIP feature distributions.} Green: healthy. Red: depressed. Violins show distributions, dots recordings, and dashed lines class means.}
\label{fig:inertia-distributions}
\end{figure}

Each row of \Cref{fig:inertia-distributions} corresponds to one cohort. Dots summarize recordings and dashed lines indicate class means. \Cref{tab:group,tab:ablation} provide the group tests and feature readouts.

\subsubsection{Elicited cohorts differ in reference-deviation consistency}
Depressed residuals concentrate at higher values in both MPDD cohorts. Healthy observations sometimes reach the same range, particularly on young. Temporal and spectral ranges overlap substantially. Elder spectral estimates can exceed one within the fixed predictor coordinates. Together, these patterns explain why reference deviation recurs as a strong feature while the remaining descriptors offer cohort-dependent additions.

On AVEC2014, the residual groups occupy a broad shared range. Depressed spectral values tend toward the lower part of the healthy range in test, but that mean contrast does not recur significantly in training or yield significant spectral-only screening. The visualization therefore complements the cross-split evidence for weak reference discrimination.

\subsubsection{Vlog overlap distinguishes mean shifts from useful combinations}
D-Vlog spectral and AR distributions overlap closely. Depressed residuals shift upward with a longer upper tail, while overlap remains. On LMVD, healthy samples extend to larger spectral-radius estimates and decay values and have a lower residual tail. These shapes explain how a residual mean difference can coexist with weak residual-only screening, and why a joint readout can benefit from the other features.

%% file: tables/design_controls.tex
\begin{table}[t]
\centering\footnotesize
\setlength{\tabcolsep}{1.7pt}
\caption{Decision and reference-construction ablations: F1 score (\%). Blue: full configurations.}
\label{tab:design-controls}
\label{tab:reference-controls}
\begin{tabular}{@{}lrrrrr@{}}
\toprule
\rowcolor{tableheader}
Configuration & D-Vlog & LMVD & \shortstack{AVEC\\2014} & \shortstack{MPDD\\elder} & \shortstack{MPDD\\young} \\
\midrule
\rowcolor{methodrow}
InertiaMPDD (full) & 98.41 & 84.98 & 66.67 & 85.71 & 80.00 \\
NSP only & 97.38 & 74.57 & 50.00 & 76.92 & 66.67 \\
w/o conditioning & 97.87 & 75.07 & 50.00 & 72.73 & 66.67 \\
Global correction & 97.35 & 80.84 & 66.67 & 72.73 & 66.67 \\
\midrule
\rowcolor{methodrow}
Full IIP (audio + visual) & 68.97 & 75.88 & 50.00 & 92.31 & 90.91 \\
Shuffled history & 67.02 & 59.68 & 35.29 & 77.78 & 90.91 \\
Repeated mean state & 71.92 & 74.81 & 46.15 & 60.00 & 63.64 \\
Visual only & 62.50 & 62.78 & 40.00 & 82.35 & 86.96 \\
Audio only & 75.68 & 64.86 & 36.36 & 77.78 & 73.68 \\
\bottomrule
\end{tabular}
\end{table}

%% file: sec/experimental_controls.tex
\subsubsection{Context and correction address different personalization steps}
\label{sec:context-controls}
Removing $z$ gives $f_{\theta_0}(s_{\le t})$. Global context replaces recording-specific conditioning with a shared training-set mean, and context swapping assigns another recording's context to the observed history. Decision correction is a separate operation: global correction replaces $b=w^\top\tilde u+c$ with $-\tau^\star$, while NSP alone sets $b=0$.

\input{tables/conditioning_supp}
On LMVD, full F1 is $84.98$, versus $75.07$ without conditioning and $80.84$ with global correction. Elder reaches $85.71$ with the full model, compared with $72.73$ for both controls. These benefits are not uniform: AVEC2014 ties global correction, and young's swapped-context result exceeds full fusion.

\subsubsection{Reference construction benefits from temporal order and multiple modalities}
\label{sec:reference-controls}
\input{tables/reference_controls_full}
\textbf{Ordered prediction benefits LMVD and elder.} Shuffling preserves the recording's states while disrupting their order, whereas repeating the mean removes within-recording changes. LMVD reaches $75.88$ F1 with full IIP, $59.68$ after shuffling and $74.81$ with a repeated mean. Elder reaches $92.31$, $77.78$ and $60.00$, respectively. Young's shuffled history reaches $90.91$, matching full IIP. Its strong static control also shows that recording-level state information can be discriminative. AVEC2014 remains weak across predictor designs.

\textbf{Learned prediction improves on simplified references.} On LMVD, full IIP reaches $75.88$ F1, compared with $70.36$ for a static MLP, $67.82$ for persistence and $65.50$ for a linear AR predictor. The corresponding elder scores are $92.31$, $66.67$, $58.82$ and $53.33$. These contrasts support learning a conditional next-state reference beyond static mapping, state repetition or linear prediction on these cohorts.

\textbf{Audio and vision provide complementary reference evidence on LMVD and MPDD.} Joint IIP reaches $75.88$ F1 on LMVD, exceeding visual-only $62.78$ and audio-only $64.86$, and exceeds either modality on both MPDD cohorts. Reference utility and the need for true temporal order therefore vary across cohorts.

\input{tables/nsp_modality_supp}
\textbf{D-Vlog has strong direct acoustic and language cues.} Adding transcripts to audiovisual NSP raises F1 from $95.74$ to $97.38$, with correction reaching $98.41$.

\subsubsection{Descriptor informativeness differs from fusion complementarity}
\label{sec:correction-inputs}
\input{tables/fusion_features_supp}
\textbf{A useful descriptor need not require the full feature combination.} $R$ alone matches full correction on AVEC2014 and exceeds it on MPDD-young, whereas the full correction is strongest among these inputs on D-Vlog and elder. Standalone informativeness and fusion complementarity are therefore distinct.

\subsubsection{Pathway weighting changes how reference evidence is used}
\input{tables/fusion_weighting_supp}
Scaling learns $\alpha$ in $\alpha m+b$. Convex fusion uses $\alpha m+(1-\alpha)q(u)$, and feature-dependent gating replaces $\alpha$ with $\alpha(u)$, where $q(u)$ is the standalone IIP score.

\textbf{Pathway balance matters when IIP is strong.} Convex fusion reaches $93.33$ F1 on elder and $100.00$ on young, compared with $85.71$ and $80.00$ for fixed additive correction. Allowing the relative pathway weights to change can therefore make better use of the strong standalone IIP scores on MPDD. Fixed additive correction remains strongest among these rules on both vlog cohorts, so a more flexible rule does not uniformly improve screening. These controls separate the informativeness of the reference descriptors from the decision rule used to combine them with NSP.

\subsubsection{D-Vlog decisions can survive simplified dynamics}
\label{sec:mechanism-diagnostics}
These single-run diagnostics hold NSP margins fixed and refit correction heads to test whether simplified references preserve the screening decisions.

\begin{table}[htbp]
\centering\small
\caption{D-Vlog diagnostic controls with a fixed population margin.}
\label{tab:diagnostic-summary}
\begin{tabular}{@{}p{0.32\linewidth}p{0.61\linewidth}@{}}
\toprule
Control & Observed relation to the diagnostic reference \\
\midrule
Zero or global context & Same test decisions as true context in one setting \\
Persistence predictor & Reproduces the conditional predictor's test decisions \\
PCA / Mahalanobis scores & Match confusion counts, each differing on two recordings \\
Global / block time shuffling & Retain screening performance \\
Repeated mean-state input & Retains strong screening performance without temporal variation \\
\bottomrule
\end{tabular}
\end{table}

Strong NSP margins make several reference variants decision-equivalent. Label-independent keyword-aligned masking of cached audiovisual intervals gives $96.84$ F1 versus $96.88$ for its diagnostic reference, testing those intervals rather than transcript-token removal from the full model.

\subsubsection{Reference coverage affects descriptor quality}
\input{tables/reference_pool_supp}
\textbf{Reference coverage affects discrimination.} Duration matching preserves strong MPDD IIP performance, whereas healthy-pool subsampling weakens LMVD and elder. The cohort-dependent response to interval masking further supports learning references that cover the available expression patterns.

%% file: tables/conditioning_supp.tex
\begin{table}[!htbp]
\centering\small
\setlength{\tabcolsep}{1.0pt}
\caption{Conditioning and correction controls: F1 score (\%).}
\label{tab:control-f1}
\begin{tabular}{@{}lrrrrr@{}}
\toprule
\rowcolor{tableheader}
Configuration & D-Vlog & LMVD & \shortstack{AVEC\\2014} & \shortstack{MPDD\\elder} & \shortstack{MPDD\\young} \\
\midrule
\rowcolor{methodrow}
InertiaMPDD & 98.41 & 84.98 & 66.67 & 85.71 & 80.00 \\
w/o conditioning & 97.87 & 75.07 & 50.00 & 72.73 & 66.67 \\
Global context & 97.87 & 75.07 & 54.55 & 72.73 & 66.67 \\
Context swapping & 97.35 & 76.55 & 33.33 & 77.78 & 90.91 \\
Global correction & 97.35 & 80.84 & 66.67 & 72.73 & 66.67 \\
\bottomrule
\end{tabular}
\end{table}

%% file: tables/reference_controls_full.tex
\begin{table}[!t]
\centering\footnotesize
\setlength{\tabcolsep}{2.0pt}
\caption{Complete IIP predictor and modality controls (\%). Blue: full IIP.}
\label{tab:reference-controls-full}
\resizebox{\linewidth}{!}{%
\begin{tabular}{@{}lrrrrrrrrrr@{}}
\toprule
\rowcolor{tableheader}
Configuration & \multicolumn{2}{c}{D-Vlog} & \multicolumn{2}{c}{LMVD} & \multicolumn{2}{c}{AVEC2014} & \multicolumn{2}{c}{MPDD-elder} & \multicolumn{2}{c}{MPDD-young} \\
\cmidrule(lr){2-3}\cmidrule(lr){4-5}\cmidrule(lr){6-7}\cmidrule(lr){8-9}\cmidrule(lr){10-11}
 & Accuracy & F1 score & Accuracy & F1 score & Accuracy & F1 score & Accuracy & F1 score & Accuracy & F1 score \\
\midrule
\rowcolor{methodrow}
Full IIP (causal Transformer) & 64.00 & 68.97 & 73.70 & 75.88 & 62.50 & 50.00 & 95.45 & 92.31 & 90.91 & 90.91 \\
Static MLP & 65.68 & 69.15 & 67.60 & 70.36 & 56.25 & 0.00 & 68.18 & 66.67 & 100.00 & 100.00 \\
Persistence & 62.72 & 63.16 & 64.10 & 67.82 & 56.25 & 46.15 & 68.18 & 58.82 & 45.45 & 33.33 \\
Linear AR predictor & 63.31 & 67.37 & 61.50 & 65.50 & 75.00 & 33.33 & 68.18 & 53.33 & 81.82 & 81.82 \\
Shuffled history & 62.72 & 67.02 & 57.20 & 59.68 & 31.25 & 35.29 & 81.82 & 77.78 & 90.91 & 90.91 \\
Repeated mean state & 66.27 & 71.92 & 73.10 & 74.81 & 56.25 & 46.15 & 63.64 & 60.00 & 63.64 & 63.64 \\
\midrule
Visual only & 53.85 & 62.50 & 60.70 & 62.78 & 25.00 & 40.00 & 86.36 & 82.35 & 86.36 & 86.96 \\
Audio only & 73.37 & 75.68 & 60.30 & 64.86 & 56.25 & 36.36 & 81.82 & 77.78 & 77.27 & 73.68 \\
\bottomrule
\end{tabular}}
\end{table}

%% file: tables/nsp_modality_supp.tex
\begin{table}[!htbp]
\centering\small
\setlength{\tabcolsep}{3.5pt}
\caption{D-Vlog NSP modality and correction controls (\%).}
\label{tab:nsp-modality}
\begin{tabular}{@{}lrr@{}}
\toprule
\rowcolor{tableheader}
Configuration & Accuracy & F1 score \\
\midrule
Visual-only NSP & 73.14 & 75.39 \\
Audio-only NSP & 92.57 & 92.82 \\
Audio--visual NSP & 95.43 & 95.74 \\
Audio--visual--text NSP & 97.14 & 97.38 \\
Audio--visual NSP + IIP & 95.43 & 95.74 \\
\rowcolor{methodrow}
InertiaMPDD & 98.29 & 98.41 \\
\bottomrule
\end{tabular}
\end{table}

%% file: tables/fusion_features_supp.tex
\begin{table}[!htbp]
\centering\small
\setlength{\tabcolsep}{1.0pt}
\caption{Fusion feature controls: F1 score (\%).}
\label{tab:fusion-features}
\begin{tabular}{@{}lrrrr@{}}
\toprule
\rowcolor{tableheader}
Correction input & D-Vlog & \shortstack{AVEC\\2014} & \shortstack{MPDD\\elder} & \shortstack{MPDD\\young} \\
\midrule
NSP only & 97.38 & 50.00 & 76.92 & 66.67 \\
NSP + $R$ & 97.89 & 66.67 & 77.78 & 85.71 \\
NSP + $\forcear$ & 97.35 & 54.55 & 66.67 & 66.67 \\
NSP + $\rho(J)$ & 97.87 & 50.00 & 72.73 & 70.00 \\
NSP + $R,\forcear$ & 97.38 & 28.57 & 77.78 & 85.71 \\
NSP + $R,\rho(J)$ & 97.89 & 28.57 & 77.78 & 85.71 \\
NSP + $\forcear,\rho(J)$ & 97.35 & 28.57 & 66.67 & 72.73 \\
\rowcolor{methodrow}
InertiaMPDD & 98.41 & 66.67 & 85.71 & 80.00 \\
\bottomrule
\end{tabular}
\end{table}

%% file: tables/fusion_weighting_supp.tex
\begin{table}[!htbp]
\centering\footnotesize
\setlength{\tabcolsep}{1.0pt}
\caption{Pathway-weighting controls: F1 score (\%).}
\label{tab:fusion-weighting}
\begin{tabular}{@{}lrrrrr@{}}
\toprule
\rowcolor{tableheader}
Decision rule & D-Vlog & LMVD & \shortstack{AVEC\\2014} & \shortstack{MPDD\\elder} & \shortstack{MPDD\\young} \\
\midrule
NSP only & 97.38 & 74.57 & 50.00 & 76.92 & 66.67 \\
IIP only & 68.97 & 75.88 & 50.00 & 92.31 & 90.91 \\
Global correction & 97.35 & 80.84 & 66.67 & 72.73 & 66.67 \\
\rowcolor{methodrow}
InertiaMPDD & 98.41 & 84.98 & 66.67 & 85.71 & 80.00 \\
Scaling & 97.87 & 75.20 & 50.00 & 48.28 & 66.67 \\
Convex fusion & 97.38 & 77.06 & 60.00 & 93.33 & 100.00 \\
\shortstack[l]{Feature-dependent\\gating} & 97.38 & 77.85 & 36.36 & 75.00 & 90.91 \\
\bottomrule
\end{tabular}
\end{table}

%% file: tables/reference_pool_supp.tex
\begin{table}[!htbp]
\centering\small
\setlength{\tabcolsep}{1.0pt}
\caption{IIP controls for duration, interval masking and reference-pool size: F1 score (\%).}
\label{tab:reference-pool}
\begin{tabular}{@{}lrrr@{}}
\toprule
\rowcolor{tableheader}
Setting & LMVD & \shortstack{MPDD\\elder} & \shortstack{MPDD\\young} \\
\midrule
IIP reference & 75.88 & 92.31 & 90.91 \\
Duration matched & 56.07 & 100.00 & 90.91 \\
AV interval masking & 80.65 & 87.50 & 85.71 \\
Healthy-pool subsampling & 70.54 & 73.68 & 95.24 \\
\bottomrule
\end{tabular}
\end{table}

%% file: sec/numerical_validation.tex
\subsection{Numerical accuracy and extraction cost}
\label{sec:numerical-validation}

\input{tables/spectral_validation}
\textbf{Thirty JVP steps balance spectral accuracy and cost.} Relative errors are $4.25$\% and $4.53$\% on D-Vlog and LMVD, with faster extraction than exact eigendecomposition. Further iterations increase runtime without consistently reducing error.

\paragraph{Incremental descriptor-extraction cost}
IIP extraction from cached backbone features takes approximately $0.23$ seconds per recording, including predictor evaluation and JVP iterations, with peak GPU memory below $1$~GiB in the recorded runs. Fitting the correction head takes approximately $0.2$--$0.3$ seconds on CPU. Backbone extraction is excluded from these measurements.

%% file: tables/spectral_validation.tex
\begin{table}[!t]
\centering\footnotesize
\setlength{\tabcolsep}{3.5pt}
\caption{Jacobian spectral estimation: numerical accuracy and extraction cost.}
\label{tab:spectral-validation}
\begin{tabular}{@{}lrrrrr@{}}
\toprule
\rowcolor{tableheader}
Cohort & Estimator & \shortstack{Mean relative\\error (\%)} & \shortstack{Spearman\\correlation} & Runtime (s) & \shortstack{Peak memory\\(GiB)} \\
\midrule
D-Vlog & Exact eigendecomposition & 0 & 1 & 0.322 & 0.059 \\
\rowcolor{methodrow}
D-Vlog & 30 JVP steps & 4.25 & 0.6086 & 0.204 & 0.058 \\
D-Vlog & 60 JVP steps & 5.32 & 0.5547 & 0.444 & 0.058 \\
D-Vlog & 120 JVP steps & 4.28 & 0.5542 & 0.761 & 0.058 \\
\midrule
LMVD & Exact eigendecomposition & 0 & 1 & 0.370 & 0.059 \\
\rowcolor{methodrow}
LMVD & 30 JVP steps & 4.53 & 0.9423 & 0.224 & 0.058 \\
LMVD & 60 JVP steps & 5.56 & 0.9221 & 0.493 & 0.058 \\
LMVD & 120 JVP steps & 4.08 & 0.9557 & 0.929 & 0.058 \\
\bottomrule
\end{tabular}
\end{table}

%% file: sec/multi_dataset_cases.tex
\subsubsection{Case Studies across Datasets}
\label{sec:multi-dataset-cases}

\begin{figure}[!tp]
\centering
\includegraphics[width=\textwidth,height=0.86\textheight,keepaspectratio]{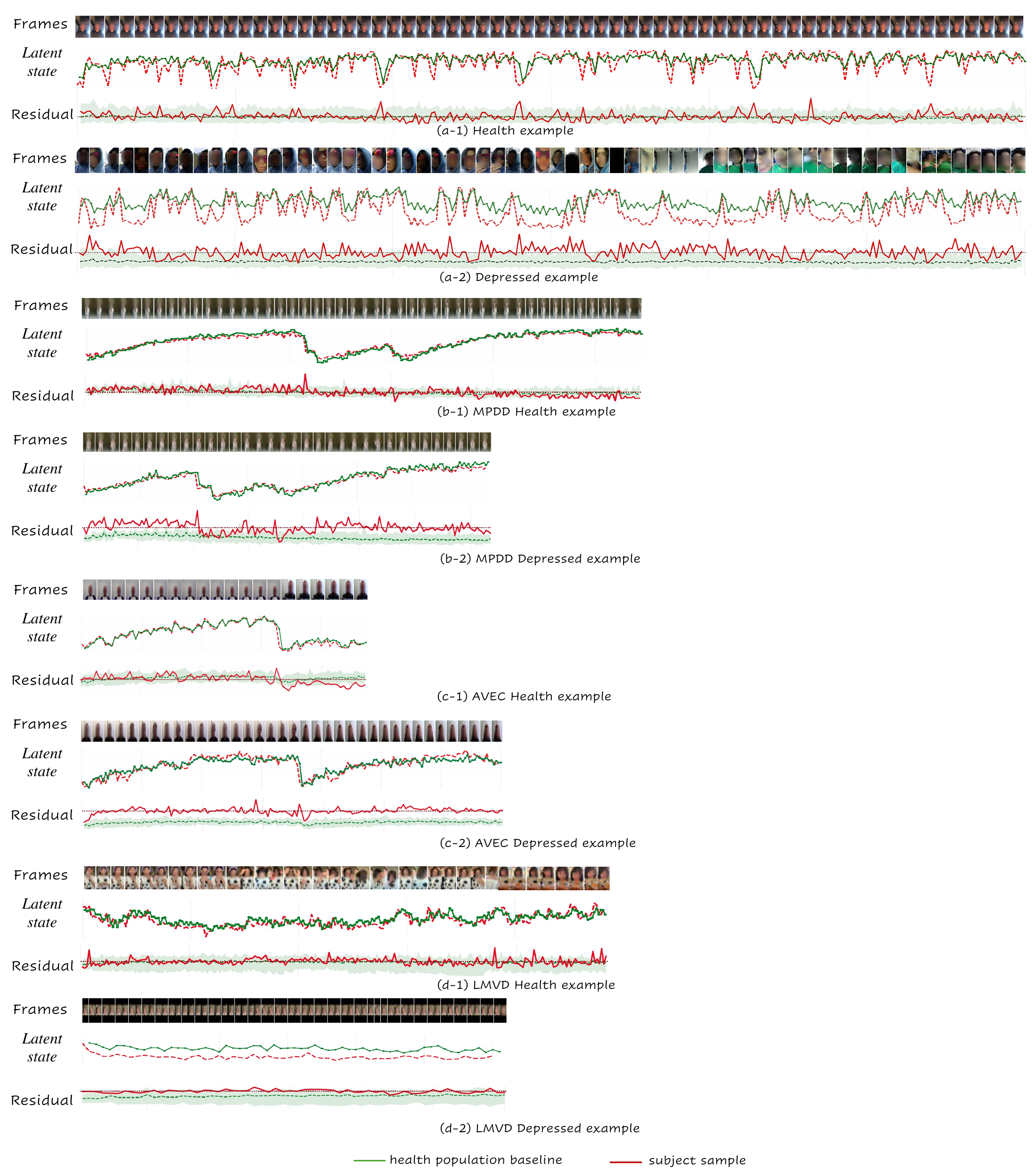}
\caption{\textbf{Case studies across datasets.} Healthy/depressed pairs from D-Vlog, MPDD-young, AVEC2014 and LMVD. Green: healthy reference. Red: subject trace or residual. Shading: healthy residual 5--95\% range.}
\label{fig:case}
\end{figure}

\paragraph{Visualization settings}
D-Vlog bands use healthy test recordings with the displayed case excluded. Pointwise 5th, 50th and 95th percentiles summarize aligned one-second traces. Its two latent curves share a first-principal-component projection fitted to the healthy pool.

The MPDD pair comes from an earlier case-study run using the official partition and contains training-partition examples. Its healthy-training reference band excludes the displayed subject. Quantitative MPDD tables use cohort-specific $80{:}20$ retraining. These cases illustrate prediction agreement under their visualization settings. Reference bands and latent projections are descriptive context for reading the trajectories.

\paragraph{Predictive agreement is more informative than visible stillness}
LMVD further separates predictive agreement from visible stillness. The healthy recording fluctuates with intermittent residual peaks, whereas the relatively smooth depressed trace remains offset from its reference. The relevant contrast is prediction agreement rather than the amount of visible motion.

\paragraph{Similar projections can conceal full-state disagreement}
In MPDD-young, the healthy trajectory follows long trends and abrupt changes with mainly local error excursions. The depressed example follows the broad projected shape but has a more persistently elevated residual. The full-state norm captures disagreement beyond the single coordinate shown in the latent curve.

\paragraph{AVEC examples explain the feature without defining the cohort}
Both AVEC2014 trajectories follow the reference's large-scale changes, while the depressed example has a residual offset. This illustrates the temporal meaning of $R$ without overriding the cohort's weak standalone separation.